\documentclass[letterpaper,10pt,conference]{ieeeconf}
\IEEEoverridecommandlockouts
\usepackage{cite}
\usepackage{graphicx}
\usepackage{amsmath,amssymb}
\usepackage{booktabs}
\usepackage{multirow}
\usepackage{array}
\usepackage{tabularx}
\usepackage{url}
\usepackage{xcolor}
\usepackage{float}
\usepackage[section]{placeins}

\newcolumntype{L}[1]{>{\raggedright\arraybackslash}p{#1}}
\title{Transforming Omnidirectional RGB-LiDAR data into 3D Gaussian Splatting}

\author{Semin Bae, Hansol Lim, Jongseong Brad Choi
\thanks{Semin Bae is with the Department of Computer Science, State University of New York, Stony Brook, NY 11794, USA (e-mail: semin.bae@stonybrook.edu). Hansol Lim, and Jongseong Brad Choi are with the Department of Mechanical Engineering, State University of New York, Stony Brook, NY 11794, USA (email: hansol.lim@stonybrook.edu; jongseong.choi@stonybrook.edu). This work supported by the National Research Foundation of Korea (NRF) grant funded by the Korea government (MSIT) (No. RS-2022-NR067080 and RS-2025-05515607). (Corresponding author: Jongseong Brad Choi).}
}

\begin{document}
\maketitle

\begin{abstract}
The demand for large-scale digital twins is rapidly growing in robotics and autonomous driving. However, constructing these environments with 3D Gaussian Splatting (3DGS) usually requires expensive, purpose-built data collection. Meanwhile, deployed platforms routinely collect extensive omnidirectional RGB and LiDAR logs, but a significant portion of these sensor data is directly discarded or strictly underutilized due to transmission constraints and the lack of scalable reuse pipelines. In this paper, we present an omnidirectional RGB-LiDAR reuse pipeline that transforms these archived logs into robust initialization assets for 3DGS. Direct conversion of such raw logs introduces practical bottlenecks: inherent non-linear distortion leads to unreliable Structure-from-Motion (SfM) tracking, and dense, unorganized LiDAR clouds cause computational overhead during 3DGS optimization. To overcome these challenges, our pipeline strategically integrates an ERP-to-cubemap conversion module for deterministic spatial anchoring, alongside PRISM---a color stratified downsampling strategy. By bridging these multi-modal inputs via Fast Point Feature Histograms (FPFH) based global registration and Iterative Closest Point (ICP), our pipeline successfully repurposes a considerable fraction of discarded data into usable SfM geometry. Furthermore, our LiDAR-reinforced initialization consistently enhances the final 3DGS rendering fidelity in structurally complex scenes compared to vision-only baselines. Ultimately, this work provides a deterministic workflow for creating simulation-grade digital twins from standard archived sensor logs.
\end{abstract}

\section{Introduction}
High-fidelity digital twins have become fundamental infrastructure for simulation, validation, and scenario generation in robotics and autonomous driving. While 3D Gaussian Splatting (3DGS) has proven highly effective for real-time view synthesis~\cite{kerbl2023}, constructing these environments traditionally relies on expensive, purpose-built data collection. Concurrently, deployed autonomous platforms routinely archive extensive volumes of omnidirectional RGB and LiDAR logs from everyday operations. Despite their high informational value, these data are predominantly underutilized due to a lack of scalable, automated reuse pipelines. Furthermore, direct RGB-only initialization from spherical imagery often suffers from geometric drift, and while LiDAR can reinforce geometric priors, adapting raw logs introduces practical bottlenecks such as excessive point-cloud density, cross-modal synchronization, and unstable alignment outcomes. Consequently, the field currently lacks a standardized protocol to bridge the gap between archived sensor logs and robust neural scene reconstruction.

To address this, we present a reproducible and auditable omnidirectional RGB-LiDAR reuse pipeline. We focus on a highly practical question: how to convert existing, underused spherical logs into robust 3DGS initialization assets with minimal manual intervention. Our pipeline explicitly transforms raw logs through a systematic sequence of stages. First, we extract keyframes by evaluating omnidirectional image overlap via ORB features and homography, subsequently matching them with LiDAR scans using predefined temporal mappings. The selected images undergo ERP-to-cubemap conversion to establish a robust spatial anchor through Structure-from-Motion (SfM) reconstruction. Concurrently, unaligned LiDAR sweeps are aggregated into a unified point cloud using Iterative Closest Point (ICP) based odometry~\cite{besl1992}. This accumulated geometry is then colorized via sensor calibration data and substantially reduced using PRISM, a color-stratified downsampling strategy~\cite{prism2026}. Finally, the downsampled LiDAR and SfM point clouds are aligned and fused using Fast Point Feature Histograms (FPFH) features~\cite{rusu2009fpfh} and ICP refinement~\cite{besl1992}, producing a comprehensive 3DGS-ready asset that integrates robust geometry, color data, and SfM points.

The primary goal of this work is to establish a deployable conversion workflow that effectively repurposes archived sensor logs. Beyond consistently improving the final 3DGS rendering fidelity, we comprehensively evaluate the system's end-to-end robustness. To this end, we report point-cloud reduction ratios, cross-modal alignment diagnostics (e.g., global fitness and ICP RMSE), and the resulting view-synthesis performance across varying initialization densities. By generating deterministic, stage-level artifacts, we offer a practical and auditable pathway for creating simulation-grade digital twins from representative real-world sensor logs.

The main contributions of this paper are as follows:
\begin{enumerate}
\item We propose a deterministic, end-to-end data-reuse pipeline that effectively repurposes archived omnidirectional RGB-LiDAR logs into robust initialization assets for 3DGS, providing explicit reuse-efficiency accounting from raw sensor streams to usable SfM geometry.
\item We establish a robust modality-bridging workflow that strategically integrates temporal synchronization, ERP-to-cubemap SfM spatial anchoring, ICP-based LiDAR aggregation, and PRISM-based color-stratified downsampling, overcoming inherent non-linear distortion and computational bottlenecks.
\item We present a comprehensive parameter sweep of the color-stratified downsampling strategy ($n \in \{1, 5, 10, 20, 50, 100\}$), providing detailed stage-level diagnostics to rigorously evaluate the robustness and limitations of cross-modal alignment.
\item We validate the proposed LiDAR-reinforced initialization against vision-only (vanilla) baselines, demonstrating consistent improvements in final 3DGS rendering fidelity in structurally complex scenes, while explicitly analyzing the quality-resource tradeoffs essential for practical digital-twin deployment.
\end{enumerate}

\section{LITERATURE REVIEW}
\subsection*{A. Neural Scene Representations and Initialization Bottlenecks}
Neural Radiance Fields (NeRF) and its multiscale variants have established high-quality, anti-aliased novel-view synthesis for complex scenes~\cite{mildenhall2020nerf,barron2021mipnerf,barron2022mipnerf360}. To overcome the slow training and rendering speeds of implicit Multi-Layer Perceptron (MLP) architectures, subsequent works introduced explicit and hybrid data structures, such as voxel grids, hash encodings, and tensor cores~\cite{sun2022dvgo,mueller2022instantngp,fridovichkeil2022plenoxels,chen2022tensorf}. Recently, 3D Gaussian Splatting (3DGS) has emerged as a highly efficient alternative, utilizing explicit anisotropic Gaussian primitives to achieve real-time rendering and rapid optimization~\cite{kerbl2023}. This explicit representation has also inspired robotics-oriented extensions for downstream perception tasks~\cite{lee2025mattgs}.

However, the explicit nature of 3DGS introduces a critical vulnerability: its optimization is heavily dependent on the quality and density of the initial point cloud, typically derived from Structure-from-Motion (SfM) pipelines~\cite{schoenberger2016sfm,schoenberger2016mvs}. Unlike implicit MLPs that can incrementally discover geometry from random initialization, 3DGS struggles to recover from unreliable or highly sparse SfM initializations, frequently resulting in severe floating artifacts or geometric collapse. Consequently, robust SfM point cloud generation has become a critical bottleneck for deploying 3DGS in large-scale or challenging autonomous driving environments.

\subsection*{B. Omnidirectional Vision and Reconstruction Challenges}
While recent neural rendering methods attempt to directly optimize radiance fields on spherical projections, they often struggle with severe non-linear distortion at the poles and uneven spatial sampling. More importantly for 3DGS, direct feature matching on raw Equirectangular Projection (ERP) imagery frequently fails in standard Structure-from-Motion (SfM) pipelines due to this inherent non-linear distortion, leading to highly sparse or inaccurate point cloud initialization~\cite{geyer2000panoramic,scaramuzza2006toolbox,usenko2018double}. To bypass this fundamental bottleneck, our pipeline incorporates an explicit ERP-to-cubemap conversion strategy. By transforming the spherical domain into six rectilinear, pinhole-equivalent faces, we enable robust feature matching, reliable camera pose tracking, and the dense point cloud initialization required for 3DGS.

\subsection*{C. LiDAR Reinforcement in 3DGS}
Vision-only 3D reconstruction fundamentally struggles with geometric ambiguities, most notably the absence of absolute metric scale and structural degradation in textureless or repetitive regions. To overcome these inherent limitations, recent works have increasingly incorporated LiDAR priors into 3DGS~\cite{lim2025lidar3dgs,chen2024lidargs,liu2025gssdf,xiao2024livgs}. Methods such as LVI-GS, GS-SDF, and LIV-GS demonstrate that fusing precise, metric depth measurements from LiDAR significantly improves the structural integrity of Gaussian primitives. Furthermore, several approaches have successfully integrated LiDAR-camera setups for robust SLAM tracking and mapping~\cite{zhang2014loam,shan2018legoloam,shan2020liosam,xu2022fastlio2}.

Despite these advancements, existing integration methods predominantly focus on tightly-coupled, online SLAM pipelines, where 3DGS is optimized on-the-fly using perfectly synchronized, sequential sensor streams. In practical autonomous driving and robotics operations, however, extensive multi-sensor data is typically processed for offline trajectories and then merely archived. While previous large-scale autonomous driving dataset literature focuses primarily on data creation, there is a stark absence of research addressing the automated conversion and reuse efficiency of these unstructured, archived logs into rendering assets. Furthermore, establishing a cross-modal alignment between these asynchronous modalities remains challenging; traditional registration techniques are prone to local minima when aligning physically distinct sensor modalities, such as scale-ambiguous, noisy SfM point clouds with dense LiDAR sweeps~\cite{rusu2009fpfh,besl1992}. Our work diverges from real-time view synthesis by proposing a deployable data-reuse pipeline that explicitly tackles cross-modal synchronization, robust global-to-local alignment (via FPFH and ICP), and the data-scale bottlenecks inherent in repurposing real-world logs.

\subsection*{D. Point Cloud Sampling and Scale Reduction}
While integrating LiDAR provides essential metric scale and structural priors, direct initialization of 3DGS with raw, accumulated LiDAR scans introduces practical computational bottlenecks. Fusing multiple sweeps generates tens of millions of points, and injecting such dense, unorganized point clouds into 3DGS leads to excessive memory footprints and over-parameterized Gaussian splitting during optimization. Traditional point cloud reduction techniques, such as uniform voxel grids or random sampling, often indiscriminately discard crucial high-frequency geometric features or minority color distributions vital for rendering fidelity.

Conversely, advanced sampling strategies like Farthest Point Sampling (FPS) or learning-based selection (e.g., PointNet++) offer superior geometric preservation but are computationally expensive when applied to large-scale archived logs~\cite{qi2017pointnetpp}. To bridge this critical gap, our data-reuse pipeline integrates a PRISM-based color-stratified downsampling strategy~\cite{prism2026}. By explicitly balancing geometric structure with calibrated RGB distributions, this approach effectively reduces point cloud density while preserving the essential multimodal priors. Ultimately, this scalable reduction enables the practical conversion of resource-intensive, discarded sensor logs into highly efficient 3DGS digital twin assets.

\section{Methodology}
Figure~\ref{fig:pipeline_overview} summarizes the proposed deterministic conversion workflow from archived omnidirectional logs to fused 3DGS initialization assets.

\setcounter{figure}{0}
\begin{figure}[H]
\centering
\includegraphics[width=0.92\linewidth]{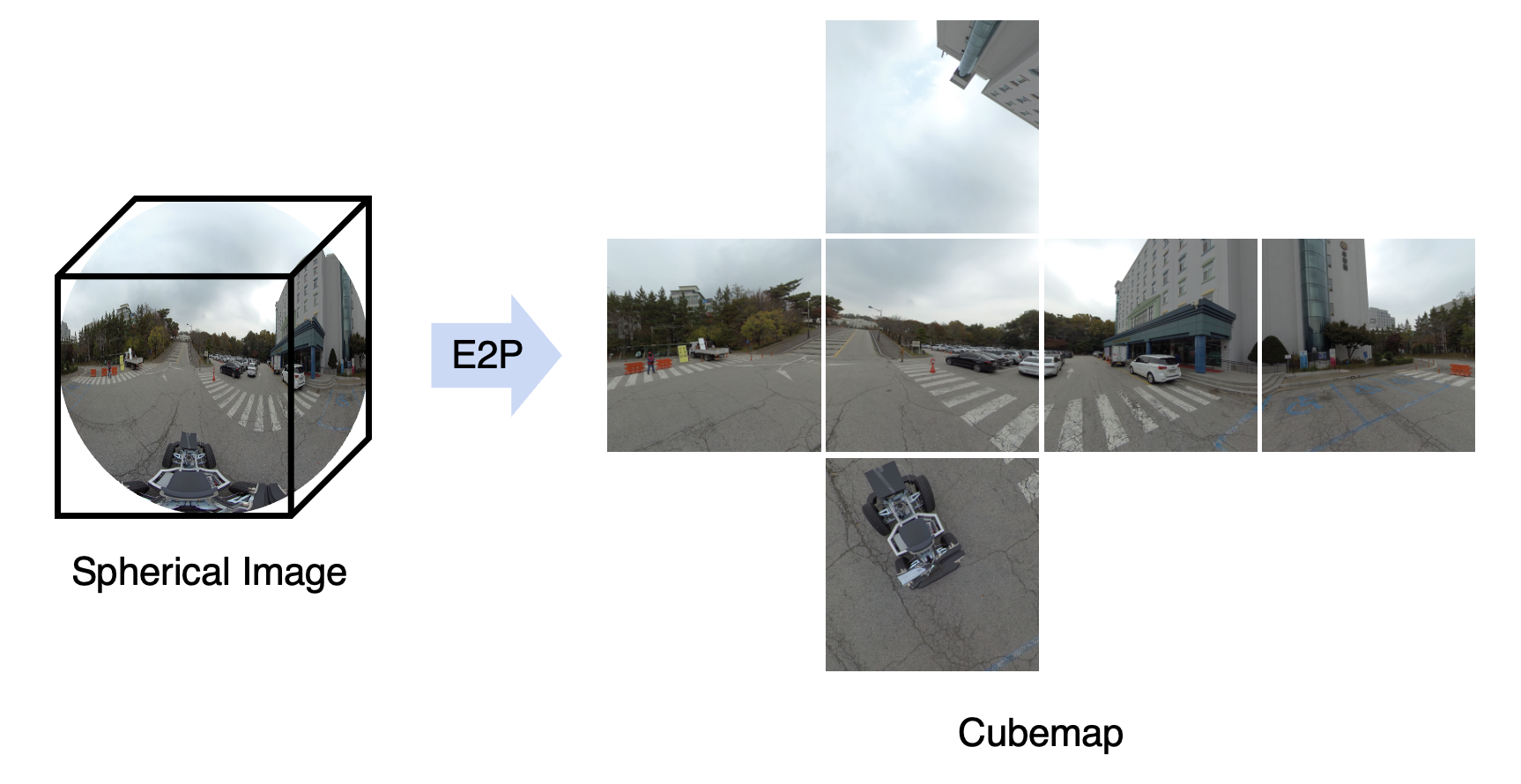}
\caption{Example of ERP-to-cubemap conversion used in our pipeline. One omnidirectional panorama is projected into six rectilinear cubemap faces for robust feature matching in SfM.}
\label{fig:erp_to_cubemap}
\end{figure}

\setcounter{figure}{1}
\begin{figure*}[!t]
\centering
\includegraphics[width=0.98\textwidth]{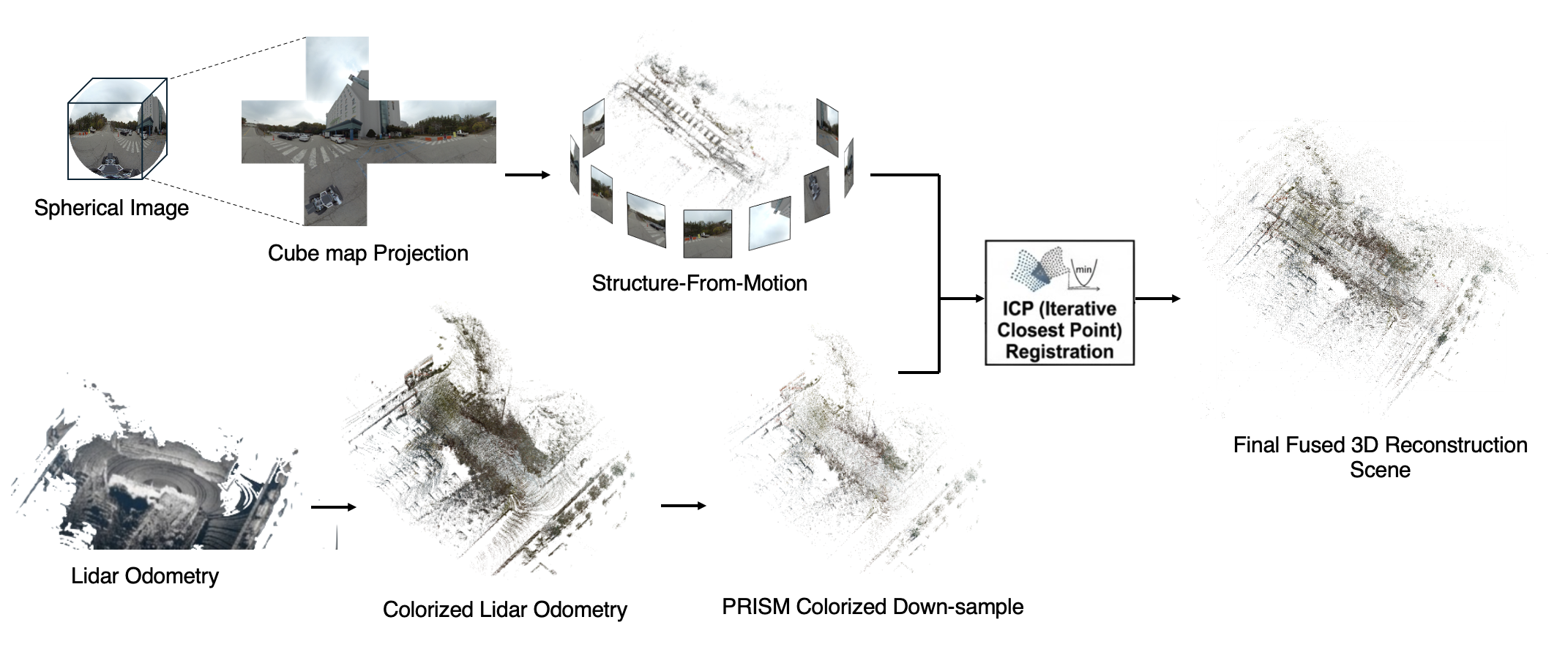}
\caption{Overview of the proposed omnidirectional RGB-LiDAR reuse pipeline. ERP images are converted to cubemaps for robust SfM anchoring, LiDAR odometry maps are colorized and PRISM-downsampled, and ICP-based registration fuses both modalities into a final 3DGS-ready asset.}
\label{fig:pipeline_overview}
\end{figure*}

\subsection*{A. Modality Bridging: ERP-to-Cubemap Projection}
Raw Equirectangular Projection (ERP) images inherently suffer from severe non-linear distortions at the poles, causing feature-matching failures in standard Structure-from-Motion (SfM). To ensure robust feature tracking, we explicitly project the ERP frames onto rectilinear cubemaps. A pixel $(u, v)$ in the ERP image corresponding to longitude $\theta \in [-\pi, \pi]$ and latitude $\phi \in [-\pi/2, \pi/2]$ is mapped to a 3D ray $\mathbf{d}$:
\begin{equation}
\mathbf{d} =
\begin{bmatrix}
\cos\phi\sin\theta \\
\sin\phi \\
\cos\phi\cos\theta
\end{bmatrix}
\label{eq:erp_ray}
\end{equation}
Intersecting this ray with a unit cube restores rectilinear constraints, allowing standard multi-view geometry pipelines to reliably extract features.
Figure~\ref{fig:erp_to_cubemap} shows an example conversion from one ERP panorama to six rectilinear cubemap faces used in our SfM stage.

\subsection*{B. Deterministic Spatial Anchoring via SfM}
Using the undistorted cubemap faces, we establish a deterministic spatial anchor. For a 3D point $\mathbf{X} \in \mathbb{R}^3$ in the world coordinate, its projection $\mathbf{u}$ onto the $i$-th camera image plane is formulated as:
\begin{equation}
s \tilde{\mathbf{u}} = \mathbf{K}_i [ \mathbf{R}_i \mid \mathbf{t}_i ] \tilde{\mathbf{X}}
\label{eq:sfm_proj}
\end{equation}
where $\mathbf{K}_i$ is the intrinsic matrix, and $\mathbf{R}_i \in SO(3)$, $\mathbf{t}_i \in \mathbb{R}^3$ denote the camera extrinsics. To prevent the non-deterministic collapse common in noisy in-the-wild logs, we strictly constrain the optimization parameters in our SfM engine, yielding a scale-ambiguous sparse point cloud $\mathcal{P}_{\text{sfm}}$ and a set of reliable camera poses.

\subsection*{C. LiDAR Colorization and PRISM Downsampling}
Directly injecting massive LiDAR scans into the 3DGS optimizer causes severe VRAM bottlenecks. Conventional downsampling (e.g., Voxel Grid) enforces spatial uniformity, severely degrading semantic and photometric priors.

To resolve this, we colorize the LiDAR points using Eq.~\eqref{eq:sfm_proj} and introduce PRISM (Color-Stratified Point Cloud Sampling). PRISM shifts the stratification domain from spatial coverage to visual complexity. Let $\mathcal{C}$ be the RGB color space partitioned into bins $\mathcal{B}_c$. By imposing a maximum point capacity $k$ per color bin, the downsampled set $\mathcal{P}_{\text{sub}}$ is dynamically generated as:
\begin{equation}
\mathcal{P}_{\text{sub}} = \bigcup_{c \in \mathcal{C}}
\operatorname{Sample}\!\left(\mathcal{B}_c,\; \min\!\left(\left|\mathcal{B}_c\right|, k\right)\right)
\label{eq:prism_sub}
\end{equation}
This explicit preservation of chromatic diversity aggressively decimates visually homogeneous geometry while retaining the texture-rich regions crucial for spherical harmonics initialization.

\subsection*{D. Robust Multi-Modal Alignment}
The scale-ambiguous $\mathcal{P}_{\text{sfm}}$ must be metrically aligned with the downsampled LiDAR cloud $\mathcal{P}_{\text{sub}}$. Given the extreme noise and sparsity of in-the-wild data, global registration frequently diverges. Therefore, we bypass exhaustive global search and utilize a robust Iterative Closest Point (ICP) optimization initialized by trajectory metadata to find the optimal rigid transformation $(\mathbf{R}^*, \mathbf{t}^*)$:
\begin{equation}
\mathbf{R}^*, \mathbf{t}^* = \arg\min_{\mathbf{R}, \mathbf{t}} \sum_{j=1}^{N} w_j
\left\| \mathbf{R} \mathbf{p}_{\text{sfm}}^{(j)} + \mathbf{t} - \mathbf{p}_{\text{sub}}^{(j)} \right\|^2
\label{eq:icp_obj}
\end{equation}
where $w_j$ is a weight function to reject dynamic outliers. This local optimization effectively bridges the coordinate gap.

\subsection*{E. 3DGS Initialization}
The aligned, multimodal point cloud is directly transformed into 3DGS initialization assets. Each point $\mathbf{p}_i \in \mathcal{P}_{\text{sub}}$ initializes the mean $\boldsymbol{\mu}_i$ of a 3D Gaussian, formulated as:
\begin{equation}
\mathcal{G}(\mathbf{x}) = \exp \left( -\frac{1}{2} (\mathbf{x} - \boldsymbol{\mu}_i)^T \boldsymbol{\Sigma}_i^{-1} (\mathbf{x} - \boldsymbol{\mu}_i) \right)
\label{eq:gaussian_init}
\end{equation}
The covariance matrix $\boldsymbol{\Sigma}_i$ is initialized based on local point density, and the projected RGB values initialize the zeroth-order spherical harmonics. The registered extrinsics from Eq.~\eqref{eq:sfm_proj} format the training views, establishing a deterministic foundation for 3DGS optimization.

\section{Experimental Setup and Implementation}
\subsection*{A. System Implementation Details}
Our data-reuse pipeline is designed to be highly memory-efficient, enabling the processing of massive multimodal sensor logs on a single workstation. All data preparation and 3D Gaussian Splatting (3DGS) rendering experiments were conducted on an NVIDIA RTX 4080 GPU with 16GB VRAM. This hardware constraint intentionally demonstrates that our PRISM-based reduction allows billion-point-scale logs to be processed without requiring enterprise-grade memory clusters. For software integration, we utilize the COLMAP framework for ERP-to-cubemap SfM sparse structure recovery and camera pose estimation~\cite{schoenberger2016sfm,schoenberger2016mvs}. Cross-modal point cloud alignment (ICP) is implemented via the Open3D library. To isolate and evaluate the pure contribution of our multimodal initialization, all rendering tests utilize the original 3DGS CUDA implementation by Kerbl et al.~\cite{kerbl2023} without any structural modifications to the training engine.

All preprocessing stages are executed with a shared configuration rather than per-scene retuning. In particular, the cubemap projection layout, SfM reconstruction policy, LiDAR color-transfer routine, and PRISM bucket sweep are kept fixed across all three sequences. This constraint is intentional: the target setting is archived-log conversion, so the pipeline is only practically useful if the same settings remain stable under different overlap patterns, texture conditions, and point-cloud densities.

\subsection*{B. Dataset and Processing Protocol}
To validate robustness and scalability across different campus trajectories, we evaluate our system on three large-scale sequences from the AIR Lab omnidirectional RGB-LiDAR dataset.

AIR Lab 360 RGB-LiDAR dataset: We utilize three unstructured campus sequences: Dormitory 1, College of Engineering, and College of Physical Edu~\cite{kim2024pair360,airlab_dataset}. This dataset features dense LiDAR paired with raw ERP (360-degree) imagery, providing an optimal stress test for our spherical-to-pinhole (ERP-to-cubemap) conversion module against severe non-linear lens distortion.

The selected sequences stress different failure modes for archived 3DGS initialization. Dormitory 1 contains repetitive facade structure and narrow road boundaries, College of Engineering introduces larger viewpoint change with mixed vegetation and vehicles, and College of Physical Edu contains broader open-space geometry with the largest accumulated clouds. Evaluating all three with the same sensor rig isolates algorithmic behavior from hardware variation and makes the later PRISM/ICP sweep directly comparable.

\subsection*{C. Baseline Configurations and PRISM Sweep}
To explicitly measure the impact of our LiDAR-reinforced initialization on 3DGS optimization, we establish the following experimental variants.

Vanilla (Baseline). A vision-only initialization trajectory utilizing only ERP keyframes, cubemap projection, and SfM sparse points, completely bypassing the LiDAR pipeline.

No-PRISM (Stress Case). A naive multimodal initialization that injects the fully densified, colorized LiDAR point cloud without downsampling. This serves as a qualitative failure case to demonstrate system collapse due to excessive memory footprint and over-parameterized Gaussian splitting.

Ours ($n$). The proposed end-to-end pipeline featuring PRISM color-stratified sampling. We conduct a parameter sweep across varying maximum points per RGB bucket, where $n \in \{1, 5, 10, 20, 30, 50, 100\}$, to analyze the tradeoff between metric registration stability and rendering fidelity.

All 3DGS variants are trained with the same camera views, iteration budget, optimizer, and renderer. The only intentional difference is the initialization asset delivered to the trainer. This matched setup is important because it turns the comparison into an initialization study rather than a broader system-level benchmark confounded by view selection or training-schedule changes.

\subsection*{D. Evaluation Metrics and Artifact Contract}
We evaluate both the efficiency of the data-reuse system and the final novel-view synthesis quality. Rendering fidelity is measured using PSNR, SSIM, and LPIPS after a fixed budget of 30,000 iterations. System-level robustness is evaluated through global alignment fitness, ICP RMSE, and the PRISM reduction ratio. Crucially, our pipeline enforces a strict artifact contract: every processing stage from LiDAR colorization to final SfM export automatically generates machine-readable JSON and CSV logs. This guarantees that the conversion of archived robot logs into 3DGS digital twins remains a fully deterministic, auditable, and reproducible engineering workflow.

\section{Results and Discussion}
\subsection{Pipeline Reuse and Throughput}
Table~\ref{tab:pipeline_stats} reports end-to-end conversion throughput and reuse proxies for the three AIR Lab sequences.
\begin{table}[!htbp]
\caption{Pipeline summary with reuse-efficiency proxies for three AirLab-360 sequences.}
\label{tab:pipeline_stats}
\centering
\setlength{\tabcolsep}{3pt}
\footnotesize
\resizebox{\columnwidth}{!}{
\begin{tabular}{lrrrrrr}
\toprule
Sequence & ERP & LiDAR & Keyframes & SfM imgs & KF reuse & SfM rec. \\
\midrule
Dormitory 1 & 280 & 280 & 103 & 509 & 36.8\% & 82.4\% \\
College of Engineering & 279 & 279 & 143 & 716 & 51.3\% & 83.4\% \\
College of Physical Edu & 479 & 479 & 170 & 907 & 35.5\% & 88.9\% \\
\bottomrule
\end{tabular}
}
\end{table}
Although the three sequences have different motion and coverage patterns, the same pipeline configuration runs without scene-specific branching. The keyframe reuse ratio (35.5\%--51.3\%) and SfM reconstruction ratio (82.4\%--88.9\%) show that a meaningful portion of archived logs can be converted into usable geometry without additional data collection.
Table~\ref{tab:pc_summary} complements this result with raw/SfM/PRISM point volumes and file sizes used for initialization.

\begin{table*}[!t]
\caption{Point-cloud data summary used for initialization diagnostics.}
\label{tab:pc_summary}
\centering
\setlength{\tabcolsep}{1.8pt}
\renewcommand{\arraystretch}{0.9}
\scriptsize
\begin{tabular*}{\textwidth}{@{\extracolsep{\fill}}llrrrrrr}
\toprule
Sequence & Metric & Raw data & SfM & $n=1$ & $n=5$ & $n=10$ & $n=20$ \\
\midrule
\multirow{2}{*}{Dormitory 1} & \# points & 2,058,126 & 82,724 & 145,437 & 426,720 & 628,446 & 878,976 \\
 & Size (MB) & 53.00 & 1.18 & 3.75 & 10.99 & 16.18 & 22.63 \\
\midrule
\multirow{2}{*}{College of Engineering} & \# points & 2,275,569 & 88,903 & 146,667 & 442,041 & 660,160 & 942,330 \\
 & Size (MB) & 58.59 & 1.27 & 3.78 & 11.38 & 17.00 & 24.26 \\
\midrule
\multirow{2}{*}{College of Physical Edu} & \# points & 3,336,973 & 248,964 & 209,707 & 600,370 & 883,179 & 1,252,454 \\
 & Size (MB) & 85.92 & 3.56 & 5.40 & 15.46 & 22.74 & 32.25 \\
\bottomrule
\end{tabular*}
\end{table*}
\renewcommand{\arraystretch}{1.0}

These volume statistics also explain why later alignment behavior is not uniform across scenes. College of Physical Edu begins with the largest accumulated LiDAR and SfM clouds, indicating strong spatial coverage but also a harder reduction problem. By contrast, Dormitory 1 enters the sweep with fewer SfM points and more repetitive structure, making it more sensitive to how aggressively PRISM balances point density against stable cross-modal correspondence.

\subsection{PRISM Sweep and Alignment Behavior}
Table~\ref{tab:prism_sweep} and Table~\ref{tab:prism_volume} summarize alignment quality and density reduction across the PRISM sweep ($n \in \{1,5,10,20,30,50,100\}$).
\begin{table*}[!t]
\caption{Alignment diagnostics grouped by $n$ (updated SLAM-colorized sweep).}
\label{tab:prism_sweep}
\centering
\setlength{\tabcolsep}{1.8pt}
\renewcommand{\arraystretch}{0.9}
\scriptsize
\begin{tabular*}{\textwidth}{@{\extracolsep{\fill}}c|ccc|ccc|ccc}
\toprule
\multirow{2}{*}{$n$} & \multicolumn{3}{c|}{Dormitory 1} & \multicolumn{3}{c|}{College of Engineering} & \multicolumn{3}{c}{College of Physical Edu} \\
\cmidrule(lr){2-4}\cmidrule(lr){5-7}\cmidrule(lr){8-10}
 & Global $\uparrow$ & ICP $\uparrow$ & RMSE $\downarrow$ & Global $\uparrow$ & ICP $\uparrow$ & RMSE $\downarrow$ & Global $\uparrow$ & ICP $\uparrow$ & RMSE $\downarrow$ \\
\midrule
1   & 0.9586 & 0.9874 & 0.3179 & 0.9944 & 0.9940 & 0.2640 & 0.9999 & 1.0000 & 0.1999 \\
5   & 0.9505 & 0.9868 & 0.3171 & 0.9926 & 0.9948 & 0.2923 & 0.9922 & 0.9974 & 0.2200 \\
10  & 0.8904 & 0.9867 & 0.3180 & 0.9127 & 0.9970 & 0.2076 & 0.9999 & 0.9999 & 0.2118 \\
20  & 0.9275 & 0.9870 & 0.3185 & 0.9978 & 0.9985 & 0.2177 & 0.9997 & 1.0000 & 0.2138 \\
30  & 0.9869 & 0.9876 & 0.2812 & 0.9941 & 0.9988 & 0.2004 & 0.9998 & 0.9983 & 0.2679 \\
50  & 0.9122 & 0.9669 & 0.3817 & 0.9980 & 0.9995 & 0.2499 & 0.9897 & 0.9992 & 0.2436 \\
100 & 0.8634 & 0.8946 & 0.3231 & 0.9923 & 0.9992 & 0.2006 & 0.9950 & 0.9999 & 0.2462 \\
\bottomrule
\end{tabular*}

\end{table*}
\renewcommand{\arraystretch}{1.0}

\begin{table*}[!t]
\caption{PRISM output size and reduction ratio grouped by $n$.}
\label{tab:prism_volume}
\centering
\setlength{\tabcolsep}{1.8pt}
\renewcommand{\arraystretch}{0.9}
\scriptsize
\begin{tabular*}{\textwidth}{@{\extracolsep{\fill}}c|cc|cc|cc}
\toprule
\multirow{2}{*}{$n$} & \multicolumn{2}{c|}{Dormitory 1} & \multicolumn{2}{c|}{College of Engineering} & \multicolumn{2}{c}{College of Physical Edu} \\
\cmidrule(lr){2-3}\cmidrule(lr){4-5}\cmidrule(lr){6-7}
 & Points & Reduction & Points & Reduction & Points & Reduction \\
\midrule
1   & 145,437   & 0.9293 & 146,667   & 0.9355 & 209,707   & 0.9372 \\
5   & 426,720   & 0.7927 & 442,041   & 0.8057 & 600,370   & 0.8201 \\
10  & 628,446   & 0.6947 & 660,160   & 0.7099 & 883,179   & 0.7353 \\
20  & 878,976   & 0.5729 & 942,330   & 0.5859 & 1,252,454 & 0.6247 \\
30  & 1,042,608 & 0.4934 & 1,130,716 & 0.5031 & 1,502,772 & 0.5497 \\
50  & 1,258,801 & 0.3884 & 1,382,223 & 0.3926 & 1,841,923 & 0.4480 \\
100 & 1,546,399 & 0.2486 & 1,723,438 & 0.2426 & 2,313,297 & 0.3068 \\
\bottomrule
\end{tabular*}
\end{table*}
\renewcommand{\arraystretch}{1.0}

As expected, larger $n$ retains more points and lowers reduction ratios. Registration quality is generally strong, but not strictly monotonic: some scenes preserve high ICP fitness while RMSE degrades at dense settings, indicating that more points do not always translate to better local alignment. Averaging across scenes reinforces the same operating-region conclusion: moderate $n$ balances compression and alignment more effectively, whereas very large $n$ mainly adds optimization cost without guaranteed downstream benefit.

Across Tables~\ref{tab:pc_summary}--\ref{tab:prism_volume}, the practical objective is not maximizing any single diagnostic in isolation but finding a density that remains robust across all stages. Low values of $n$ offer strong compression and lighter downstream optimization, yet they can underrepresent the color diversity needed for stable Gaussian seeding. High values retain more structure, but once local correspondence becomes noisy the extra points mainly increase computational load instead of producing better rendering.

\subsection{3DGS Quantitative Comparison}
Table~\ref{tab:gs_absolute} and Table~\ref{tab:gs_delta} report absolute outcomes and deltas against the vanilla baseline after 30k iterations.
\begin{table*}[!t]
\caption{Absolute 3DGS metrics at 30k iterations.}
\label{tab:gs_absolute}
\centering
\setlength{\tabcolsep}{3pt}
\scriptsize
\resizebox{\textwidth}{!}{
\begin{tabular}{llrrrrrr}
\toprule
Sequence & Variant & PSNR $\uparrow$ & SSIM $\uparrow$ & LPIPS $\downarrow$ & Train sec & Points & Model MB \\
\midrule
Dormitory 1 & Vanilla & 27.4389 & 0.7817 & 0.3541 & 1030 & 1,205,789 & 1338 \\
Dormitory 1 & Ours ($n=5$) & 27.5551 & 0.7816 & 0.3536 & 1256 & 1,207,071 & 1326 \\
Dormitory 1 & Ours ($n=50$) & 27.7505 & 0.7854 & 0.3486 & 1611 & 1,542,370 & 1404 \\
Dormitory 1 & Ours ($n=100$) & 27.8043 & 0.7864 & 0.3473 & 1717 & 1,692,739 & 1439 \\
College of Engineering & Vanilla & 27.6077 & 0.7773 & 0.3538 & 626 & 1,054,880 & 1149 \\
College of Engineering & Ours ($n=5$) & 27.7042 & 0.7747 & 0.3597 & 774 & 1,033,338 & 1141 \\
College of Engineering & Ours ($n=50$) & 27.8964 & 0.7793 & 0.3526 & 1053 & 1,375,343 & 1218 \\
College of Engineering & Ours ($n=100$) & 27.9105 & 0.7798 & 0.3515 & 1147 & 1,514,791 & 1252 \\
College of Physical Edu & Vanilla & 26.4412 & 0.7205 & 0.3672 & 700 & 1,219,200 & 1598 \\
College of Physical Edu & Ours ($n=5$) & 26.3555 & 0.7018 & 0.3864 & 809 & 1,076,663 & 1515 \\
College of Physical Edu & Ours ($n=50$) & 26.4927 & 0.7101 & 0.3783 & 1086 & 1,518,619 & 1634 \\
College of Physical Edu & Ours ($n=100$) & 26.5048 & 0.7108 & 0.3769 & 1188 & 1,716,351 & 1679 \\
\bottomrule
\end{tabular}
}
\end{table*}

The LiDAR-initialized variants are clearly $n$-sensitive. Higher-density settings ($n=50,100$) consistently improve PSNR, while SSIM/LPIPS gains vary by scene. This pattern indicates that LiDAR priors are beneficial when cross-modal alignment quality is sufficiently reliable.

\begin{figure*}[!t]
\centering
\includegraphics[width=0.78\textwidth]{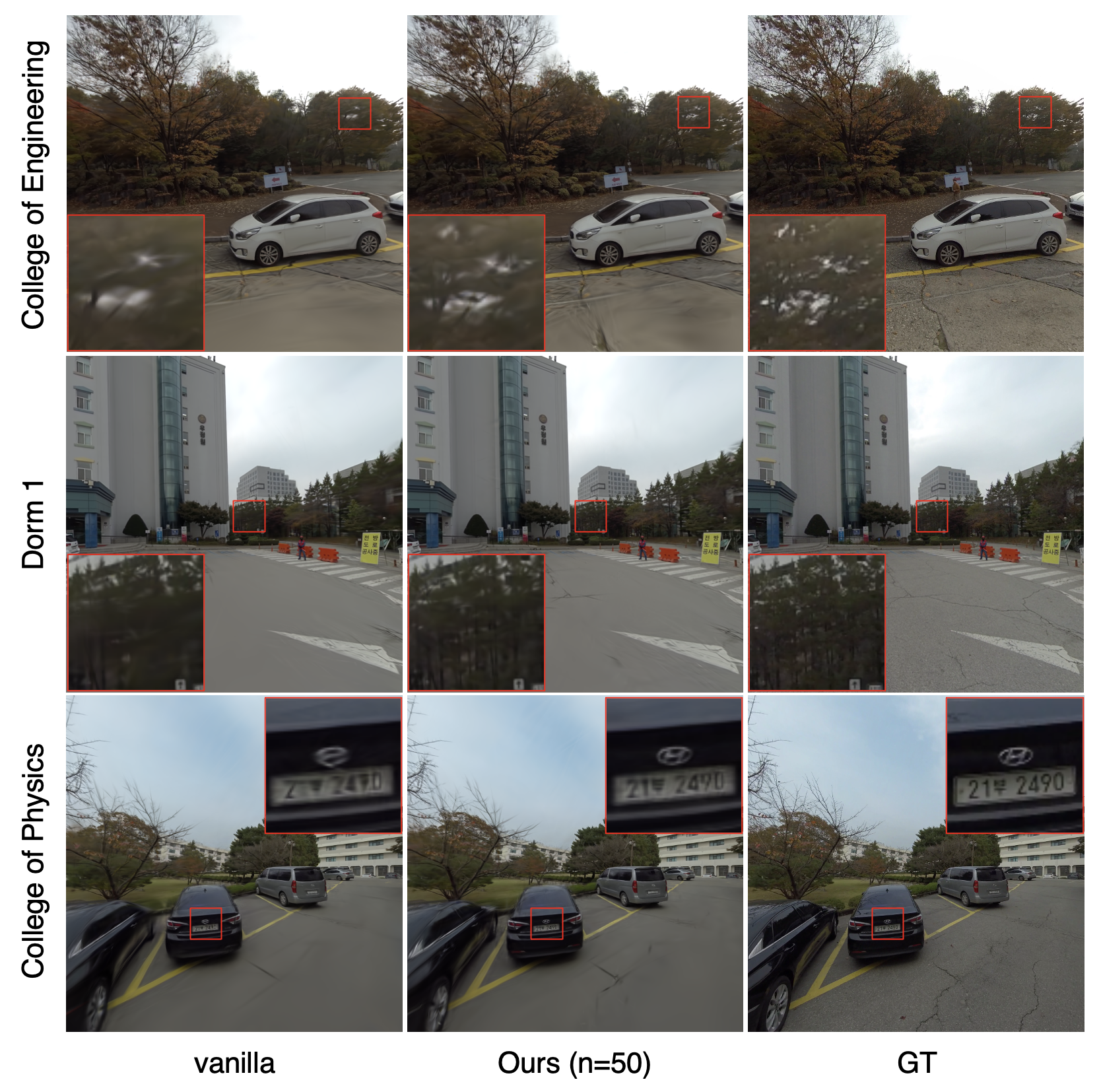}
\caption{Qualitative comparison across the three sequences. Columns show Vanilla, Ours ($n=50$), and GT.}
\label{fig:gs_qual}
\end{figure*}

Resource cost follows the same direction as initialization density. Denser LiDAR priors increase training time, retained Gaussian count, and model size because the optimizer starts from a richer geometric prior and preserves more structure during refinement. The important practical result is that these overheads remain within a single-workstation budget, so the gains at $n=50$ and $n=100$ are achieved without assuming specialized multi-GPU infrastructure. Figure~\ref{fig:gs_qual} shows the same trend qualitatively, with LiDAR-initialized runs recovering sharper boundaries and clearer local detail in thin branches and plate-like text.

\begin{table}[!htbp]
\caption{Delta metrics (LiDAR-initialized variant - vanilla) at 30k iterations.}
\label{tab:gs_delta}
\centering
\setlength{\tabcolsep}{4pt}
\footnotesize
\resizebox{\columnwidth}{!}{
\begin{tabular}{llrrr}
\toprule
Sequence & Variant & $\Delta$PSNR & $\Delta$SSIM & $\Delta$LPIPS \\
\midrule
Dormitory 1 & $n=5$ & +0.1162 & -0.0001 & -0.0005 \\
Dormitory 1 & $n=50$ & +0.3116 & +0.0036 & -0.0056 \\
Dormitory 1 & $n=100$ & +0.3654 & +0.0047 & -0.0068 \\
College of Engineering & $n=5$ & +0.0966 & -0.0026 & +0.0059 \\
College of Engineering & $n=50$ & +0.2887 & +0.0020 & -0.0012 \\
College of Engineering & $n=100$ & +0.3028 & +0.0025 & -0.0022 \\
College of Physical Edu & $n=5$ & -0.0857 & -0.0187 & +0.0192 \\
College of Physical Edu & $n=50$ & +0.0516 & -0.0105 & +0.0112 \\
College of Physical Edu & $n=100$ & +0.0636 & -0.0097 & +0.0097 \\
\bottomrule
\end{tabular}
}
\end{table}

\subsection{Discussion}
Dormitory 1 and College of Engineering benefit more consistently from denser LiDAR priors, while College of Physical Edu remains harder due to broad open-space geometry and weaker fine-scale consistency. Across all three scenes, the main bottleneck is not pipeline executability but alignment reliability: global and ICP fitness can remain high even when rendering gains plateau, showing that correspondence quality matters more than raw point count alone.

For practical deployment, LiDAR reinforcement should be treated as conditional rather than universal. We recommend gating training on minimum registration quality, checking RGB-LiDAR projection overlays before 3DGS optimization, and retaining a vanilla fallback because the better initialization remains scene-dependent. All major claims are tied to machine-generated CSV/JSON artifacts under identical training budgets, so the reported metrics should be interpreted as deterministic initialization diagnostics rather than full held-out trajectory benchmarks.

\subsection{Limitations and Future Work}
This study has four main limitations. First, residual spherical-image distortion still limits robust RGB color-to-LiDAR matching during LiDAR colorization in difficult regions. Second, experiments were conducted on only three outdoor sequences from the same sensor platform, so broader validation across more scenes, devices, and environmental conditions is still required. Third, we did not perform exhaustive parameter search for PRISM and ICP, so the current settings may be suboptimal for some trajectories. Fourth, the evaluation is offline and mostly static-scene oriented, leaving dynamic-object robustness and real-time deployment constraints for future work.

\section{Conclusion}
We presented a deterministic omnidirectional RGB-LiDAR reuse pipeline that converts archived sensor logs into 3DGS-ready initialization assets. The reported reuse ratios and alignment/quality diagnostics show that substantial portions of previously underused logs can be repurposed into practical digital-twin inputs. Results across three real-world sequences further show that LiDAR reinforcement is effective but alignment-dependent, making robust cross-modal registration the key factor for consistent gains. Overall, this work provides an auditable baseline workflow for scalable digital-twin construction from existing field data.

\section*{ACKNOWLEDGMENT}
This work was supported by the National Research Foundation of Korea (NRF) grant funded by the Korea government (MSIT) (No. RS-2022-NR067080 and RS-2025-05515607).


{\scriptsize
\begin{thebibliography}{10}
\setlength{\itemsep}{0pt}
\setlength{\parsep}{0pt}
\setlength{\parskip}{0pt}
\providecommand{\url}[1]{#1}
\csname url@samestyle\endcsname
\providecommand{\newblock}{\relax}
\providecommand{\bibinfo}[2]{#2}
\providecommand{\BIBentrySTDinterwordspacing}{\spaceskip=0pt\relax}
\providecommand{\BIBentryALTinterwordstretchfactor}{4}
\providecommand{\BIBentryALTinterwordspacing}{\spaceskip=\fontdimen2\font plus
\BIBentryALTinterwordstretchfactor\fontdimen3\font minus
  \fontdimen4\font\relax}
\providecommand{\BIBforeignlanguage}[2]{{%
\expandafter\ifx\csname l@#1\endcsname\relax
\typeout{** WARNING: IEEEtran.bst: No hyphenation pattern has been}%
\typeout{** loaded for the language `#1'. Using the pattern for}%
\typeout{** the default language instead.}%
\else
\language=\csname l@#1\endcsname
\fi
#2}}
\providecommand{\BIBdecl}{\relax}
\BIBdecl

\bibitem{kerbl2023}
B.~Kerbl, G.~Kopanas, T.~Leimkuhler, and G.~Drettakis, ``3d gaussian splatting
  for real-time radiance field rendering,'' \emph{ACM Transactions on
  Graphics}, vol.~42, no.~4, 2023.

\bibitem{besl1992}
P.~J. Besl and N.~D. McKay, ``A method for registration of 3-d shapes,''
  \emph{IEEE Transactions on Pattern Analysis and Machine Intelligence},
  vol.~14, no.~2, pp. 239--256, 1992.

\bibitem{prism2026}
H.~Lim, M.~Im, and J.~B. Choi, ``Prism: Color-stratified point cloud
  sampling,'' \emph{arXiv preprint arXiv:2601.06839}, 2026.

\bibitem{rusu2009fpfh}
R.~B. Rusu, N.~Blodow, and M.~Beetz, ``Fast point feature histograms (fpfh) for
  3d registration,'' in \emph{ICRA}, 2009.

\bibitem{mildenhall2020nerf}
B.~Mildenhall, P.~P. Srinivasan, M.~Tancik, J.~T. Barron, R.~Ramamoorthi, and
  R.~Ng, ``Nerf: Representing scenes as neural radiance fields for view
  synthesis,'' \emph{European Conference on Computer Vision (ECCV)}, 2020.

\bibitem{barron2021mipnerf}
J.~T. Barron, B.~Mildenhall, D.~Verbin, P.~P. Srinivasan, and P.~Hedman,
  ``Mip-nerf: A multiscale representation for anti-aliasing neural radiance
  fields,'' in \emph{ICCV}, 2021.

\bibitem{barron2022mipnerf360}
J.~T. Barron, B.~Mildenhall, M.~Tancik, P.~Hedman, R.~Martin-Brualla, and P.~P.
  Srinivasan, ``Mip-nerf 360: Unbounded anti-aliased neural radiance fields,''
  in \emph{CVPR}, 2022.

\bibitem{sun2022dvgo}
C.~Sun, M.~Sun, and H.-T. Chen, ``Direct voxel grid optimization: Super-fast
  convergence for radiance fields reconstruction,'' \emph{arXiv preprint
  arXiv:2111.11215}, 2022.

\bibitem{mueller2022instantngp}
T.~Mueller, A.~Evans, C.~Schied, and A.~Keller, ``Instant neural graphics
  primitives with a multiresolution hash encoding,'' \emph{ACM Transactions on
  Graphics}, vol.~41, no.~4, 2022.

\bibitem{fridovichkeil2022plenoxels}
S.~Fridovich-Keil, A.~Yu, M.~Chen, M.~Tancik, B.~Recht, and A.~Kanazawa,
  ``Plenoxels: Radiance fields without neural networks,'' in \emph{CVPR}, 2022.

\bibitem{chen2022tensorf}
A.~Chen, Z.~Xu, A.~Geiger, J.~Yu, and H.~Su, ``Tensorf: Tensorial radiance
  fields,'' in \emph{ECCV}, 2022.

\bibitem{lee2025mattgs}
J.~W. Lee, H.~Lim, S.~Yang, and J.~B. Choi, ``Matt-gs: Masked attention-based
  3dgs for robot perception and object detection,'' \emph{arXiv preprint
  arXiv:2503.19330}, 2025.

\bibitem{schoenberger2016sfm}
J.~L. Schoenberger and J.-M. Frahm, ``Structure-from-motion revisited,'' in
  \emph{CVPR}, 2016.

\bibitem{schoenberger2016mvs}
J.~L. Schoenberger, E.~Zheng, M.~Pollefeys, and J.-M. Frahm, ``Pixelwise view
  selection for unstructured multi-view stereo,'' in \emph{ECCV}, 2016.

\bibitem{geyer2000panoramic}
C.~Geyer and K.~Daniilidis, ``A unifying theory for central panoramic systems
  and practical implications,'' in \emph{ECCV}, 2000.

\bibitem{scaramuzza2006toolbox}
D.~Scaramuzza, A.~Martinelli, and R.~Siegwart, ``A toolbox for easily
  calibrating omnidirectional cameras,'' in \emph{IROS}, 2006.

\bibitem{usenko2018double}
V.~Usenko, N.~Demmel, and D.~Cremers, ``The double sphere camera model,'' in
  \emph{3DV}, 2018.

\bibitem{lim2025lidar3dgs}
H.~Lim, H.~Chang, J.~B. Choi, and C.~M. Yeum, ``Lidar-3dgs: Lidar reinforcement
  for multimodal initialization of 3d gaussian splats,'' \emph{Computers and
  Graphics}, vol. 132, p. 104293, 2025.

\bibitem{chen2024lidargs}
H.~Zhao, W.~Guan, and P.~Lu, ``Lvi-gs: Tightly-coupled lidar-visual-inertial
  slam using 3d gaussian splatting,'' \emph{arXiv preprint arXiv:2411.02703},
  2024.

\bibitem{liu2025gssdf}
J.~Liu, Y.~Wan, B.~Wang, C.~Zheng, J.~Lin, and F.~Zhang, ``Gs-sdf:
  Lidar-augmented gaussian splatting and neural sdf for geometrically
  consistent rendering and reconstruction,'' \emph{arXiv preprint
  arXiv:2503.10170}, 2025.

\bibitem{xiao2024livgs}
R.~Xiao, W.~Liu, Y.~Chen, and L.~Hu, ``Liv-gs: Lidar-vision integration for 3d
  gaussian splatting slam in outdoor environments,'' \emph{arXiv preprint
  arXiv:2411.12185}, 2024.

\bibitem{zhang2014loam}
J.~Zhang and S.~Singh, ``Loam: Lidar odometry and mapping in real-time,'' in
  \emph{Robotics: Science and Systems (RSS)}, 2014.

\bibitem{shan2018legoloam}
T.~Shan and B.~Englot, ``Lego-loam: Lightweight and ground-optimized lidar
  odometry and mapping on variable terrain,'' in \emph{IROS}, 2018.

\bibitem{shan2020liosam}
T.~Shan, B.~Englot, C.~Ratti, and D.~Rus, ``Lio-sam: Tightly-coupled lidar
  inertial odometry via smoothing and mapping,'' in \emph{IROS}, 2020.

\bibitem{xu2022fastlio2}
W.~Xu, Y.~Cai, D.~He, J.~Lin, and F.~Zhang, ``Fast-lio2: Fast direct
  lidar-inertial odometry,'' \emph{IEEE Transactions on Robotics}, vol.~38,
  no.~4, pp. 2053--2073, 2022.

\bibitem{qi2017pointnetpp}
C.~R. Qi, L.~Yi, H.~Su, and L.~J. Guibas, ``Pointnet++: Deep hierarchical
  feature learning on point sets in a metric space,'' in \emph{NeurIPS}, 2017.

\bibitem{kim2024pair360}
G.~Kim, D.~Son, S.~Bae, K.~Kim, Y.~Jeon, S.~Lee, S.~Kim, S.~Kim, J.~Choi,
  J.~Kwak, J.~Choi, and J.~Paik, ``Pair360: A paired dataset of high-resolution
  360\textdegree{} panoramic images and lidar scans,'' \emph{IEEE Robotics and
  Automation Letters}, vol.~9, no.~11, pp. 9550--9557, 2024.

\bibitem{airlab_dataset}
{Advanced Intelligence and Robotics Laboratory (AIR Lab)}, ``Air lab 360
  rgb-lidar dataset portal,''
  \url{https://airlabkhu.github.io/PAIR-360-Dataset/}, 2024, accessed:
  2026-03-05.

\end{thebibliography}
}
\end{document}